\documentclass[a4paper,twoside]{article}

\usepackage{epsfig}
\usepackage{calc}
\usepackage{cite}
\usepackage{makecell}
\usepackage{subfig}
\usepackage{amssymb}
\usepackage{url}
\usepackage{amstext}
\usepackage{amsmath}
\usepackage{amsthm}
\usepackage{multicol}
\usepackage{pslatex}
\usepackage{apalike}
\usepackage[bottom]{footmisc}
\usepackage{placeins}
\usepackage{float}
\usepackage{SCITEPRESS}     

\begin{document}

\title{Combining Metric Learning and Attention Heads For Accurate and Efficient Multilabel Image Classification}

\author{\authorname{Kirill Prokofiev\orcidAuthor{0000-0001-9619-0248}, Vladislav Sovrasov\orcidAuthor{0000-0001-6525-2602}}
\affiliation{Intel, Munich, Germany}
\email{\{kirill.prokofiev, vladislav.sovrasov\}@intel.com}
}


\abstract{
  Multi-label image classification allows predicting a set of labels
  from a given image. Unlike multiclass classification, where
  only one label per image is assigned, such a setup is applicable for a broader range of applications.
  In this work we revisit two popular approaches to multilabel classification: transformer-based heads
  and labels relations information graph processing branches. Although transformer-based heads are considered to achieve
  better results than graph-based branches, we argue that with the proper training strategy, graph-based methods can demonstrate
  just a small accuracy drop, while spending less computational resources on inference.
  In our training strategy, instead of Asymmetric Loss (ASL), which is the de-facto
  standard for multilabel classification, we introduce its metric learning modification.
  In each binary classification sub-problem it operates with $L_2$ normalized feature vectors coming from a backbone and enforces angles
  between the normalized representations of positive and negative samples to be as large as possible.
  This results in providing a better discrimination ability, than binary cross entropy loss does on unnormalized features.
  With the proposed loss and training strategy, we obtain SOTA results among single modality methods on widespread multilabel classification
  benchmarks such as MS-COCO, PASCAL-VOC, NUS-Wide and Visual Genome 500.
  Source code of our method is available as a part of the OpenVINO{\texttrademark} Training
  Extensions\footnote{\url{https://github.com/openvinotoolkit/deep-object-reid/tree/multilabel}}.
}

\onecolumn \maketitle \normalsize \setcounter{footnote}{0} \vfill

\section{\uppercase{Introduction}}

Starting from the impressive AlexNet \cite{alexnet} breakthrough on the ImageNet benchmark \cite{imagenet_cvpr09},
the deep-learning era has drastically changed approaches to almost every computer vision task.
Throughout this process multiclass classification problem was a polygon for developing new
architectures \cite{He2016DeepRL,Howard2019SearchingFM,Tan2019EfficientNetRM} and learning paradigms \cite{chen2020simple,NEURIPS2020_d89a66c7,he2019moco}.
At the same time, multilabel classification has been developing not so intensively, although the presence of several
labels on one image is more natural, than having one hard label. Due to a lack of specialized multilabel datasets, researchers
turned general object detection datasets such as MS-COCO \cite{Lin2014MicrosoftCC} and PASCAL VOC \cite{Everingham2009ThePV} into
challenging multilabel classification benchmarks by removing bounding boxes from the data annotation
and leveraging only their class labels.

Despite the recent progress in solving the mentioned benchmarks, the latest works focus mainly on the resulting model accuracy,
not taking into account computational complexity \cite{Liu2021Query2LabelAS} or the use of outdated
training techniques \cite{Chen2019MultiLabelIR}, introducing promising model architectures at the same time.

In this work, we are revisiting the latest approaches to multilabel classification, to propose
a lightweight solution suitable for real-time applications and also improve the performance-accuracy tradeoff of the
existing models.

The main contributions of this paper are as follows:
\begin{itemize}
  \item We proposed a modification of ML-GCN \cite{Chen2019MultiLabelIR} that adds graph attention operations \cite{velickovic2018graph} and performs graph and
  CNN features fusion in a more conventional way than generating a set of binary classifiers in the graph branch.
  \item  We demonstrated that using a proper training strategy, one can decrease the performance gap between
  transformer-based heads and label co occurrence modeling via graph attention.
  \item We first applied the metric learning paradigm to multilabel a classification task and
  proposed a modified version of angular margin binary loss \cite{Wen2021SphereFace2BC}, which adds an ASL \cite{Baruch2021AsymmetricLF} mechanism to it.
  \item We verified the effectiveness of our loss and overall training strategy with comprehensive experiments
  on widespread multilabel classification benchmarks: PASCAL VOC, MS-COCO, Visual Genome \cite{Krishna2016VisualGC} and NUS-WIDE \cite{Chua2009NUSWIDEAR}.
\end{itemize}

\section{\uppercase{Related Work}}

Historically, multilabel classification was attracting less attention than the multiclass scenario,
but nonetheless there is still great progress in that field. Notable progress was achieved by
developing advanced loss functions \cite{Baruch2021AsymmetricLF}, label co occurrence modeling \cite{Chen2019MultiLabelIR,Yuan2022GraphAT},
designing advanced classification heads \cite{Liu2021Query2LabelAS,ridnik2021mldecoder,resAttn} and discovering
architectures taking into account spatial distribution of objects via exploring attentional regions
 \cite{Wang2017MultilabelIR,Gao2021LearningTD}.

Conventionally, a multilabel classification task is transformed into a set of
binary classification tasks, which are solved by optimizing a binary cross-entropy loss function.
Each single-class classification subtask suffers from a hard positives-negatives imbalance.
The more classes the training dataset contains, the more negatives we have in each of the single-class subtasks, because
an image typically contains a tiny fraction of the vast number of all classes. A modified asymmetric loss \cite{Baruch2021AsymmetricLF},
that down weights and hard-thresholds easy negative samples, showed impressive results, reaching state-of-the-art results on multiple popular
multi-label datasets without any sophisticated architecture tricks. These results indicate that a proper choice of a loss
function is crucial for multilabel classification performance.

Another promising direction is designing class-specific classifiers instead of using a fully connected
layer on top of a single feature vector produced by a backbone network. This approach
also doesn't introduce additional training steps and marginally increases model complexity.
Authors of \cite{resAttn} propose a drop-in replacement of the global average pooling layer that
generates class-specific features for every category. Leveraging compact transformer heads for
generating such features \cite{Liu2021Query2LabelAS,ridnik2021mldecoder} turned out to be even more
effective. This approach assumes pooling class-specific features by employing learnable embedding queries.

Considering the distribution of objects locations, or statistical label relationships requires
data pre-processing and additional assumptions \cite{Chen2019MultiLabelIR,Yuan2022GraphAT} or
sophisticated model architecture \cite{Wang2017MultilabelIR,Gao2021LearningTD}.
For instance, \cite{Chen2019MultiLabelIR,Yuan2022GraphAT} represents labels by word embeddings; then a directed graph is built over these label
representations, where each node denotes a label. Then stacked GCNs are learned over this graph to obtain a
set of object classifiers. The method relies on an ability to represent labels as words, which is not always possible.
Spatial distribution modeling requires placing a RCNN-like \cite{Girshick2014RichFH}
module inside the model \cite{Wang2017MultilabelIR,Gao2021LearningTD}, which drastically increases
the complexity of the training pipeline.

\section{\uppercase{Method}}

In this section, we describe the overall training pipeline and details of our approach.
We aimed not only to achieve competitive results, but also to make the training more
friendly to the end user and adaptive to data. Thus, following the principles described in \cite{prokofiefSovrasov2021},
we use lightweight model architectures, hyperparameters optimization and early stopping.

\begin{figure*}[t!]
  \centering
  \includegraphics[width=0.6\textwidth]{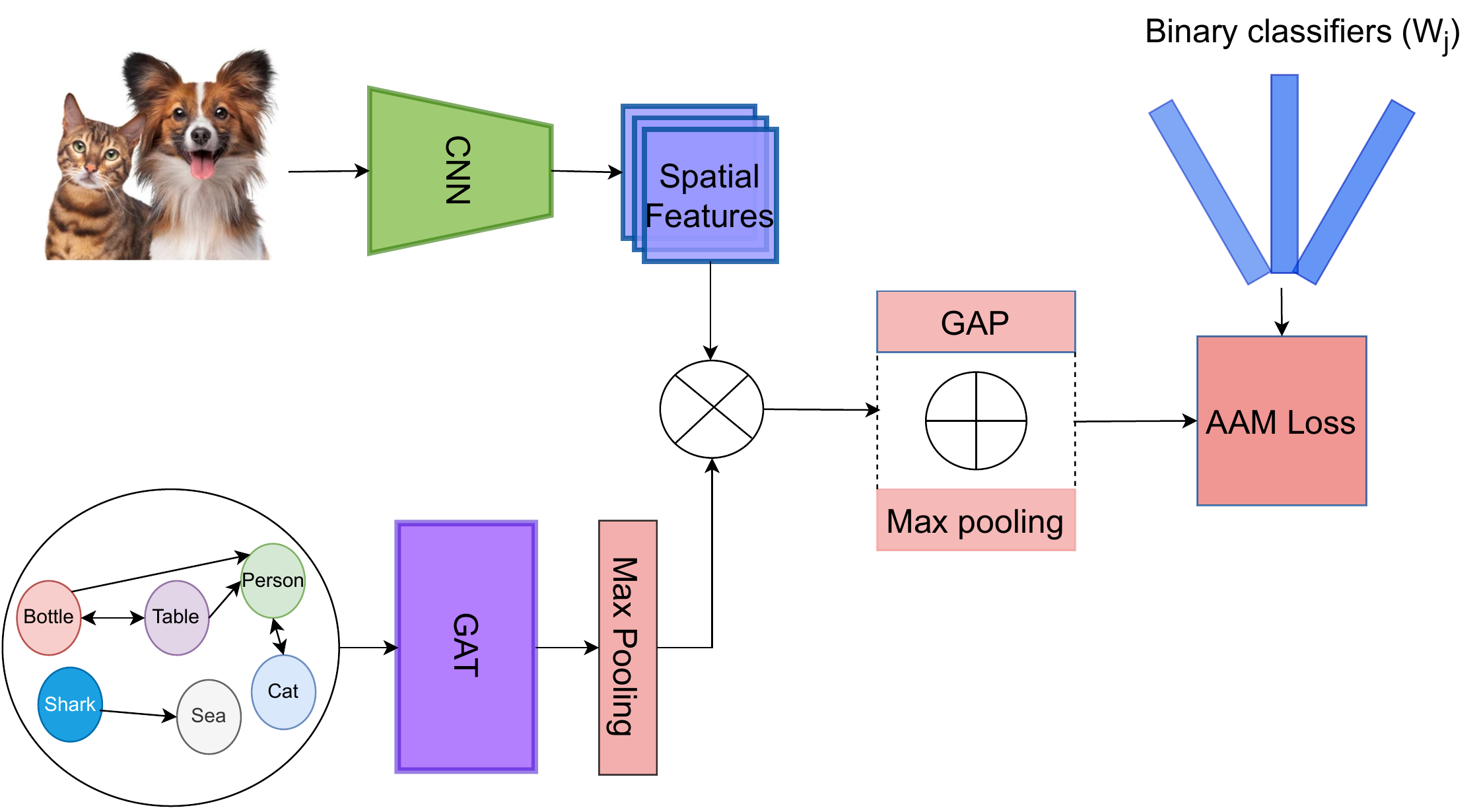}
  \caption{Overall diagram of the proposed GAT-based feature reweighting approach. Features obtained from the label relations graph by applying GAT are
  used to reweight channels in the CNN spatial features. Later, a pooling operation is applied to the reweighted features to obtain a vector representation. Finally, the resulting vectors
  are fed into binary classifiers $W_j$ and optimized with the AAM loss, which is introduced in this work.}
  \label{fig:re-weighting_scheme}
\end{figure*}

\subsection{Model architecture}

We chose EfficientNetV2 \cite{Tan2021EfficientNetV2SM} and TResNet \cite{Ridnik2021TResNetHP}
as base architectures for performing multilabel image classification.
Namely, we conducted all the experiments on TResNet-L, EfficientNetV2 small, and large.
On top of these backbones, we used two different features aggregation approaches
and compared their effectiveness and performance.

\subsection{Transformer multilabel classification head}
As a representer of transformer-based feature aggregation methods,
we use the ML-Decoder \cite{ridnik2021mldecoder} head. It provides up to $K$ feature vectors
(where $K$ is the number of classes) as a model output instead of a single class-agnostic vector
when using a standard global average pooling (GAP) head. Let's denote $x\in \mathbb{R}^{C\times H\times W}$
as a model input, then the model $F$ with parameters $W$ produces a downscaled multi-channel
featuremap $f=F_W(x)\in\mathbb{R}^{S\times \frac{H}{d}\times\frac{W}{d}}$, where $S$ is the number of output channels,
$d$ is the spatial downscale factor. That featuremap is then
passed to the ML-Decoder head: $v=MLD(f) \in \mathbb{R}^{M \times L}$, where $M$ is the embedding dimension,
$L \le K$ is the number of groups in decoder. Finally, vectors $v$ are projected to $K$ class logits by a
fully-connected (if $L=K$) or group fully-connected projection (if $L<K$) as it is described in \cite{ridnik2021mldecoder}.
In our experiments we set $L=\min(100,K)$. Also, we $L_2$ normalize the arguments of all dot products
in projections in case if we need to attach a metric learning loss to the ML-Decoder head.

\subsection{Graph attention multilabel branch}
The original structure of the graph processing branch from \cite{Chen2019MultiLabelIR} supposes generating classifiers
right in this branch and then applying them directly to the features generated by a backbone.
This approach is incompatible with the transformer-based head or any other processing of the
raw spatial features $f$, like CSRA \cite{resAttn}. To alleviate this limitation, we propose
the architecture shown in Figure \ref{fig:re-weighting_scheme}.

First, we generate the label correlation matrix $Z\in\mathbb{R}^{K \times K}$ in the same way as in \cite{Chen2019MultiLabelIR}.
Together with the word embeddings  $G\in\mathbb{R}^{K \times N}$, obtained with GLOVE \cite{Makarenkov2016LanguageMW} model,
where $N=300$ is word embeddings dimension, we utilize $Z$
as an input of the Graph Attention Network \cite{velickovic2018graph}. Unlike \cite{Yuan2022GraphAT},
we use estimations of conditional probabilities to build $Z$, rather than fully relying on GLOVE and calculating cosine similarities.

We process the input with the graph attention layers and obtain output $h\in\mathbb{R}^{S \times K}$.
Then we derive the most influential features through the max pooling operation and receive the
weights $w\in\mathbb{R}^{S}$ for further re-weighting of the CNN spatial features: $\tilde{f} = w \bigodot f$.
Next, we apply global average pooling and max-pooling operations to $\tilde f$ in parallel,
sum the results and obtain the final latent embedding $\tilde{v}\in\mathbb{R}^{S}$.
The embedding $\tilde{v}$ is finally passed to the binary classifiers. Instead of applying
a simple spatial pooling, we can pass the weighted features $\tilde f$ to the ML-Decoder or
any other features processing module.

The main advantage of using the graph attention (GAT) branch for features re-weighting
over the transformer head is a tiny computational and model complexity overhead at the inference stage.
Since the GAT branch has the same input for any image, we can compute the result of
its execution only once, before starting the inference of the resulting model.
At the same time, the GAT requires a vector representation of labels. Such
representations can be generated by a text-to-vec model in case we have a meaningful
description for all labels (even single word ones). This condition doesn't always hold:
some datasets could have untitled labels. How to generate representations for labels in that case is
still an open question.

\subsection{Angular margin binary classification}
\label{sec:aam_loss}
Recently, asymmetric loss \cite{Baruch2021AsymmetricLF} has become a standard loss option
for performing multilabel classification. By design it penalizes each logit with a modified
binary cross-entropy loss. Asymmetric handling of positives and negatives allows ASL
to down-weight the negative part of the loss to tackle the positives-negatives imbalance problem.
But this approach leaves a room for improvement from the model's discriminative ability perspective.

Angular margin losses are known for generating more discriminative classification features
than the cross-entropy loss, which is a must-have property for recognition tasks \cite{AMSoftmax,Wen2021SphereFace2BC,Sovrasov2021BuildingCE}.

\begin{figure}[t!]
  \centering
  \subfloat[Loss for $p_-$]{{\includegraphics[width=0.4\textwidth]{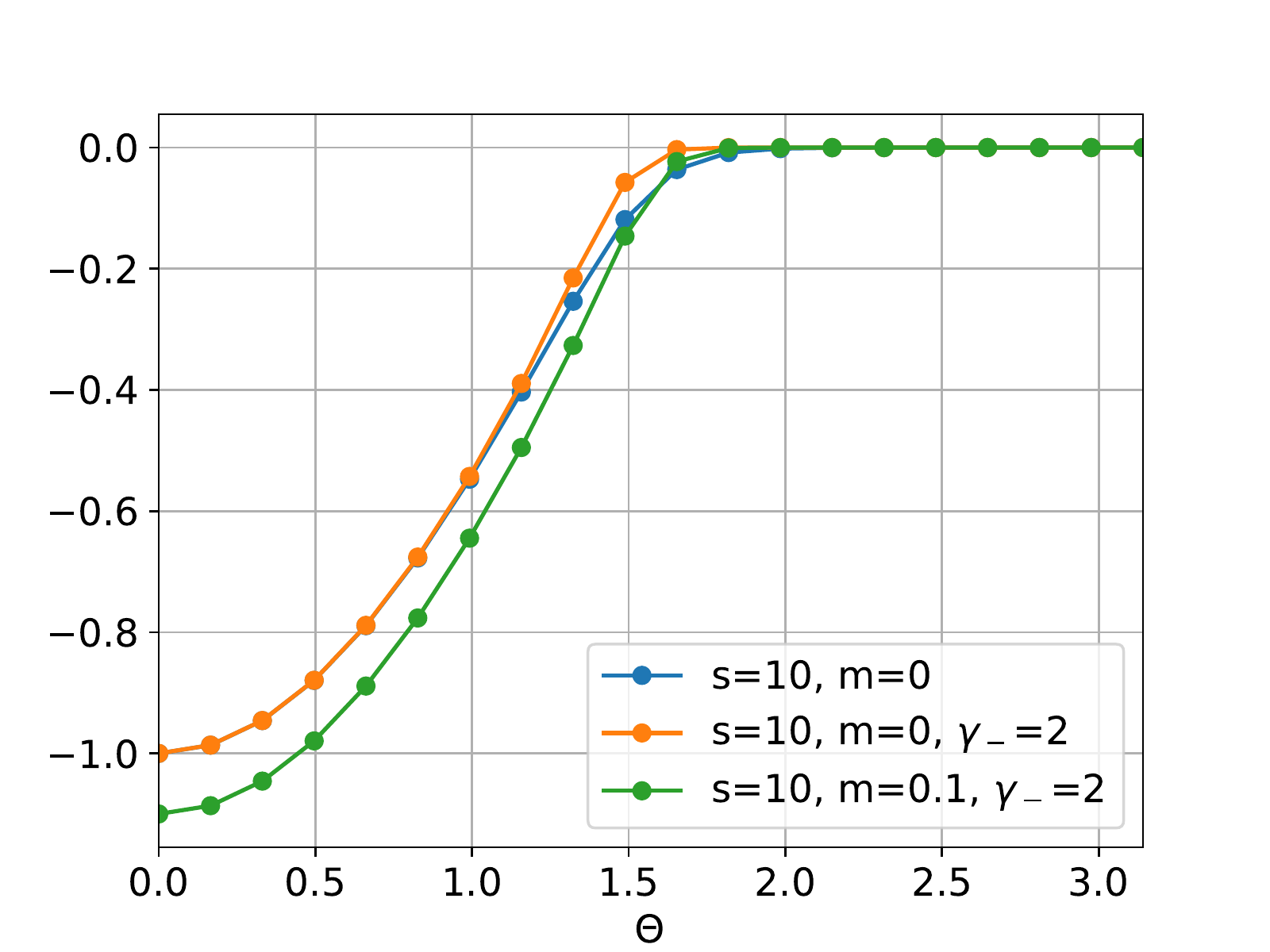} }}
  \qquad
  \subfloat[Loss for $p_+$]{{\includegraphics[width=0.4\textwidth]{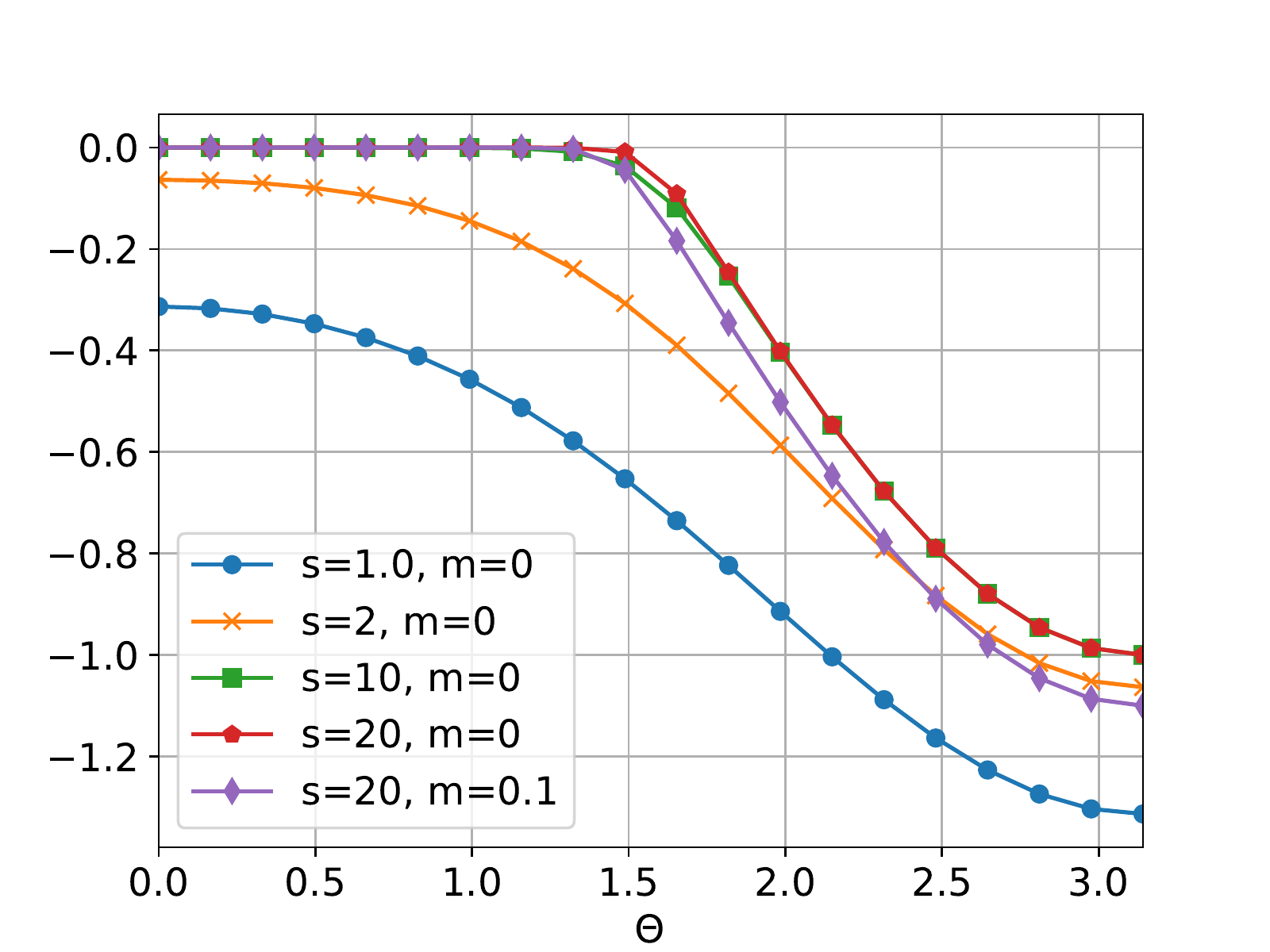} }}
  \caption{Plots of positive and negative parts of AAM with varying hyperparameters}
  \label{fig:loss}
\end{figure}

We propose joining paradigms from \cite{Baruch2021AsymmetricLF} and \cite{Wen2021SphereFace2BC} to build even stronger loss for multilabel classification.
Denote the result of the dot product between the normalized class embedding $v_j$ produced by ML-Decoder
($v_j=\tilde{v},\:j=\overline{1,K}$ if we use a backbone alone or with an optional GAT-based head) and the $j$-th binary
classifier $W_j$ as $cos\Theta_j$. Then, for a training sample $x$ and corresponding embeddings set $v$
we formulate our asymmetric angular margin loss (AAM) as:

\begin{displaymath}
  L_{AAM}(v,y)=-\sum_{j=1}^K L_j (cos\Theta_j,y)
\end{displaymath}
\begin{equation}
\label{eq:aam}
  L_j (cos\Theta_j,y) = \frac{k}{s}y p_-^{\gamma^-}\log{p_+} + \frac{1-k}{s}(1-y)p_+^{\gamma^+}\log{p_-}, \\
\end{equation}
\begin{align*}
  p_+=\sigma{(s(cos\Theta_j-m))}, \\
  p_-=\sigma{(-s(cos\Theta_j+m))},
\end{align*}
where $s$ is a scale parameter, $m$ is an angular margin, $k$ is negative-positive weighting coefficient,
$\gamma^+$ and $\gamma^-$ are weighting parameters from ASL. Despite the large number of hyperparameters,
some of them could be safely fixed (like $\gamma^+$ and $\gamma^-$ from ASL). The effect of varying $s$ saturates
when increasing $s$ (see Figure \ref{fig:loss}b), and if the suitable value
of this parameter is large enough, we don't need to tune it precisely.
Also, values of $m$ should be close to $0$, because it duplicates to some extent the effect of $s$ and $\gamma$
and can even bring undesirable increase of the negative part of AAM (see Figure \ref{fig:loss}a).
Detailed analysis of the hyperparameters is provided in Section \ref{sec:ablation}.

\subsection{Details of the training strategy}
\label{sec:training_strategy}

As in our former work \cite{prokofiefSovrasov2021}, we aim to make the training pipeline for multilabel
classification reliable, fast and adaptive to dataset, so we use the following components:
\begin{itemize}
  \item SAM \cite{Foret2020SharpnessAwareMF} optimizer with no bias decay \cite{bagOfTricks}
  is a default optimizer;
  \item EMA weights averaging for an additional protection from overfitting;
  \item Initial learning rate estimation process from \cite{prokofiefSovrasov2021};
  \item OneCycle \cite{Smith2018ADA} learning rate scheduler;
  \item Early stopping heuristic: if the best result on the validation subset hasn't been improved during 5 epochs, and
  evaluation results stay below the EMA averaged sequence of the previous best results, the training process stops;
  \item Random flip, pre-defined Randaugment \cite{NEURIPS2020_d85b63ef} strategy and Cutout \cite{Devries2017ImprovedRO} data augmentations.
\end{itemize}

\section{\uppercase{Experiments}}
In this section, we conduct a comparison of our approach with the state-of-the-art solutions on popular multilabel benchmarks.
Besides the mAP metric, we present the GFLOPs for each method to consider the speed/accuracy trade-off.
Also, we show the results of the ablation study to reveal the importance of each training pipeline component.

\subsection{Datasets}
To compare the performance of different methods, we picked up widespread datasets for multilabel classification
that are listed in Table \ref{tab:data}. According to the table, all of them have a noticeable
positives-negatives imbalance on each image: the average number of presented labels per image is significantly
lower than the number of classes.

Also, we use a precisely annotated subset of OpenImages V6 \cite{Kuznetsova2018DatasetV4} for pre-training purposes.
We join the original training and testing parts into one training subset and use the original validation subset for
testing.

\begin{table*}
  \caption{Image classification datasets which were used for training.}
  \label{tab:data}
  \centering
  \begin{tabular}{l|c|c|c|c}
    Dataset & \# of classes & \multicolumn{2}{c}{\# of Images} & Avg \# of  \\
            &                   & Train     &   Validation & labels per image \\ \hline
    Pascal VOC 2007 \cite{Everingham2009ThePV} & 20 & 5011 & 4952 & 1.58 \\
    MS-COCO 2014 \cite{Lin2014MicrosoftCC} & 80 & 117266 & 4952 & 2.92 \\
    NUS-WIDE \cite{Chua2009NUSWIDEAR} & 81 & 119103 & 50720 & 2.43 \\
    Visual Genome 500 \cite{Krishna2016VisualGC} & 500 & 82904 & 10000 & 13.61 \\ \hline
    OpenImages V6 \cite{Kuznetsova2018DatasetV4} & 601 & 1866950 & 41151 & 5.23 \\
  \end{tabular}
\end{table*}

\subsection{Evaluation protocol}

We adopt commonly used metrics for evaluation of multilabel classification models:
mean average precision (mAP) over all categories, overall precision (OP), recall (OR), F1-measure (OF1) and per category
precision (CP), recall (CR), F1-measure (CF1). We use the mAP as a main metric; others are provided
when an advanced comparison of approaches is conducted.
In every operation where confidence thresholding is required, threshold 0.5 is substituted.
The exact formulas of the mentioned metrics can be found in \cite{Liu2021Query2LabelAS}.

\subsection{Pretraining}

Although the multilabel classification task is similar to multiclass classification, a multilabel model should
pay attention to several objects on an image, instead of concentrating its attention on one object.
Thus, additional task-specific pre-training on a large-scale dataset looks beneficial to standard ImageNet
pre-training, and it was shown in recent works \cite{ridnik2021mldecoder}. In this work we utilize OpenImages V6
for pretraining. According to our experiments, the precisely-annotated subset containing 1.8M images is enough to get a
substantial improvement over the ImageNet weights.

To obtain pretrained weights on OpenImages for our models, we use the ML-Decoder head, $224\times224$ input resolution,
ASL loss with $\gamma^+=0$ and $\gamma^-=7$, learning rate $0.001$, and 50 epochs of training with OneCycle scheduler.

For TResNet-L we use the weights provided with the source code in \cite{ridnik2021mldecoder}.

\subsection{Comparison with the state-of-the-art}

In all cases where we use EfficientNet-V2-s as a backbone we also use our training strategy from Section \ref{sec:training_strategy} for a fair comparison.
For ASL loss \cite{Baruch2021AsymmetricLF} we set $lr=0.0001$, $\gamma^-=4$, $\gamma^+=0$ as suggested in the original paper \cite{Baruch2021AsymmetricLF}.
We share the following hyperparameters across all the experiments: $m=0.0$, $k=0.7$, EMA decay factor equals to $0.9997$, $\rho=0.05$ for the SAM optimizer.
Other hyperperameters for particular datasets were found empirically or via coarse grid search.

In Table \ref{tab:results_coco} results on MS-COCO are presented. For this dataset we set
$s=23$, $lr=0.007$, $\gamma^-=1$, $\gamma^+=0$.
With our AAM loss, we can achieve a state-of-the-art result using TResNet-L as a backbone.
At the same time, the combination of EfficientNetV2-s with ML-Decoder and AAM loss outperforms TResNet-L with
ASL, while consuming 3.5x less FLOPS. GCN/GAT branches perform slightly worse than ML-Decoder, but
still improves results over EfficientNetV2-s + ASL with a marginal computational cost at inference.

\begin{table*}[h!]
  \caption{Comparison with the state-of-the-art on MS-COCO dataset.}
  \label{tab:results_coco}
  \centering
  \begin{tabular}{l|c|c|c|c}
    Method & Backbone & Input resolution & mAP & GFLOPs\\ \hline
    ASL \cite{Baruch2021AsymmetricLF}  & TResNet-L & 448x448 & 88.40 & 43.50 \\
    Q2L \cite{Liu2021Query2LabelAS} & TResNet-L & 448x448 & 89.20 & 60.40 \\
    GATN \cite{Yuan2022GraphAT} & ResNeXt-101 & 448x448 & 89.30 & 36.00 \\
    Q2L \cite{Liu2021Query2LabelAS} & TResNet-L & 640x640 & 90.30 & 119.69 \\
    ML-Decoder \cite{ridnik2021mldecoder} & TResNet-L & 448x448 & 90.00 & 36.15 \\
    ML-Decoder \cite{ridnik2021mldecoder} & TResNet-L & 640x640 & 91.10 & 73.42 \\
    \hline
    ASL$^*$ & EfficientNet-V2-s & 448x448 & 87.05 & 10.83 \\
    ML-GCN$^*$ & EfficientNet-V2-s & 448x448 & 87.50 & 10.83 \\
    Q2L$^*$ & EfficientNet-V2-s & 448x448 & 87.35 & 16.25 \\
    ML-Decoder$^*$ & EfficientNet-V2-s & 448x448 & 88.25 & 12.28 \\
    \hline
    GAT re-weighting (ours) & EfficientNet-V2-s & 448x448 & 87.70 & \textbf{10.83} \\
    GAT re-weighting (ours) & TResNet-L & 448x448 & 89.95 & 35.20 \\
    ML-Decoder + AAM (ours) & EfficientNet-V2-s & 448x448 & 88.75 & 12.28 \\
    ML-Decoder + AAM (ours) & EfficientNet-V2-L & 448x448 & 90.10 & 49.92 \\
    ML-Decoder + AAM (ours) & TResNet-L & 448x448 & \textbf{90.30} & 36.15 \\
    ML-Decoder + AAM (ours) & TResNet-L & 640x640 & \textbf{91.30} & 73.42 \\
    \multicolumn{5}{l}{\footnotesize{$^*$Trained by us using our training strategy}}\\
  \end{tabular}
\end{table*}

Results on Pascal-VOC can be found in Table \ref{tab:results_voc}. We set $s=17$, $lr=0.005$, $\gamma^-=2$, $\gamma^+=1$
to train our models on this dataset. Our modification of the GAT branch outperforms
ML-Decoder when using EfficientNet-V2-s, while AAM loss gives a small performance boost and allows achieving SOTA with TResNet-L.
Also, on Pascal-VOC EfficientNet-V2-s with all of the considered additional graph branches or heads demonstrates a great speed/accuracy trade-off outperforming
TResNet-L with ASL.

\begin{table*}[h!]
  \caption{Comparison with the state-of-the-art on Pascal-VOC dataset.}
  \label{tab:results_voc}
  \centering
  \begin{tabular}{l|c|c|c|c}
    Method & Backbone & Input resolution & mAP & GFLOPs\\ \hline
    ASL \cite{Baruch2021AsymmetricLF} & TResNet-L & 448x448 & 94.60 & 43.50 \\
    Q2L \cite{Liu2021Query2LabelAS} & TResNet-L & 448x448 & 96.10 & 57.43 \\
    GATN \cite{Yuan2022GraphAT} & ResNeXt-101 & 448x448 & 96.30 & 36.00 \\
    ML-Decoder \cite{ridnik2021mldecoder} & TResNet-L & 448x448 & 96.60 & 35.78 \\
    \hline
    ASL$^*$ & EfficientNet-V2-s & 448x448 & 94.24 & 10.83 \\
    Q2L$^*$ & EfficientNet-V2-s & 448x448 & 94.94 & 15.40 \\
    ML-GCN$^*$ & EfficientNet-V2-s & 448x448 & 95.25 & 10.83 \\
    ML-Decoder$^*$ & EfficientNet-V2-s & 448x448 & 95.54 & 12.00 \\
    \hline
    GAT re-weighting (ours) & EfficientNet-V2-s & 448x448 & \textbf{96.00} & \textbf{10.83} \\
    GAT re-weighting (ours) & TResNet-L & 448x448 & 96.67 & 35.20 \\
    ML-Decoder + AAM (ours) & EfficientNet-V2-s & 448x448 & 95.86 & 12.00 \\
    ML-Decoder + AAM (ours) & EfficientNet-V2-L & 448x448 & 96.05 & 49.92 \\
    ML-Decoder + AAM (ours) & TResNet-L & 448x448 & \textbf{96.70} & 35.78 \\
    \multicolumn{5}{l}{\footnotesize{$^*$Trained by us using our training strategy}}
  \end{tabular}
\end{table*}

Tables \ref{tab:NUS_results} and \ref{tab:VG500_results} show results on NUS and VG500 datasets.
For NUS dataset we set $s=23$, $lr=0.009$, $\gamma^-=2$, $\gamma^+=1$. For VG500 the hyperparameters are $s=25$, $lr=0.005$, $\gamma^-=1$, $\gamma^+=0$.
ML-Decoder clearly outperforms GCN and GAT branches on NUS-WIDE dataset. Also on NUS EfficientNet-V2-s with the AAM loss
and GAT or ML-decoder head performs significantly better than the original implementations of Q2L and ASL with TResNet-L.

On VG500 we don’t provide results of applying GCN or GAT branches, because this dataset
has unnamed labels, so we can not apply a text-to-vec model to generate representations of graph nodes.
Applying AAM with ML-Decoder on VG500 together with increased resolution allows achieving
SOTA performance on this dataset as well.

As a result of the experiments, we can conclude that the combination of ML-Decoder with AAM loss gives
an optimal performance on all of the considered datasets, while using the GAT-based branch
could lead to better inference speed at the price of a small accuracy drop. In particular,
EfficientNetV2-s model with the GAT-based head or ML-Decoder on top outperforms T-ResNet-L
with ASL, while consuming at least 3.5x less FLOPS. These results enable real-time applications
of high-accuracy multilabel classification models on edge devices with a small computational budget.
At the same time, applying AAM loss, ML-Decoder and carefully designed training strategy allows achieving SOTA-level
accuracy with TResNet-L.

\begin{table*}[h!]
  \caption{Comparison with the state-of-the-art on NUS-WIDE.}
  \label{tab:NUS_results}
  \centering
  \begin{tabular}{l|c|c|c|c}
    Method & Backbone & Input resolution & mAP & GFLOPs\\ \hline
    GATN \cite{Yuan2022GraphAT} & ResNeXt-101 & 448x448 & 59.80 & 36.00 \\
    ASL \cite{Baruch2021AsymmetricLF} & TResNet-L & 448x448 & 65.20 & 43.50 \\
    Q2L \cite{Liu2021Query2LabelAS} & TResNet-L & 448x448 & 66.30 & 60.40 \\
    \hline
    ML-GCN$^*$ & EfficientNet-V2-s & 448x448 & 66.30 & 10.83 \\
    ASL$^*$ & EfficientNet-V2-s & 448x448 & 65.20 & 10.83 \\
    Q2L$^*$ & EfficientNet-V2-s & 448x448 & 65.79 & 16.25 \\
    ML-Decoder$^*$ & EfficientNet-V2-s & 448x448 & 67.07 & 12.28 \\
    \hline
    GAT re-weighting (ours) & EfficientNet-V2-s & 448x448 & 66.85 & \textbf{10.83} \\
    GAT re-weighting (ours) & TResNet-L & 448x448 & 68.10 & 35.20 \\
    ML-Decoder + AAM (ours) & EfficientNet-V2-s & 448x448 & 67.60 & 12.28 \\
    ML-Decoder + AAM (ours) & TResNet-L & 448x448 & \textbf{68.30} & 36.16 \\
    \multicolumn{5}{l}{\footnotesize{$^*$Trained by us using our training strategy}}
  \end{tabular}
\end{table*}

\begin{table*}[h!]
  \caption{Comparison with the state-of-the-art on VG500.}
  \label{tab:VG500_results}
  \centering
  \begin{tabular}{l|c|c|c|c}
    Method & Backbone & Input resolution & mAP & GFLOPs\\ \hline
    C-Tran \cite{Lanchantin2021GeneralMI} & ResNet101 & 576x576 & 38.40 & - \\
    Q2L \cite{Liu2021Query2LabelAS} & TResNet-L & 512x512 & 42.50 & 119.37 \\
    \hline
    ASL$^*$ & EfficientNet-V2-s & 576x576 & 38.84 & 17.90 \\
    Q2L$^*$ & EfficientNet-V2-s & 576x576 & 40.35 & 32.81 \\
    ML-Decoder$^*$ & EfficientNet-V2-s & 576x576 & 41.20 & 20.16 \\
    \hline
    ML-Decoder + AAM (ours) & EfficientNet-V2-s & 576x576 & 42.00 & 20.16 \\
    ML-Decoder + AAM (ours) & TResNet-L & 576x576 & \textbf{43.10} & 59.63 \\
    \multicolumn{5}{l}{\footnotesize{$^*$Trained by us using our training strategy}}
  \end{tabular}
\end{table*}

\subsection{Ablation study}
\label{sec:ablation}

To demonstrate the impact of each component on the whole pipeline, we add them to a baseline one by one.
As a baseline we take EfficientNetV2-s backbone and ASL loss with SGD optimizer.
We set all the hyper parameters of the ASL loss and learning rate as in \cite{Baruch2021AsymmetricLF}.
We use the training strategy described in Section \ref{sec:training_strategy} for all the experiments.

In Table \ref{tab:comp_ablation} we can see that each component brings an improvement, except adding the GAT branch.
ML Decoder has enough capacity to learn labels correlation information, so further cues, that provide the GAT branch,
don’t improve the result. Also, we can see that tuning of the $\gamma$ parameters is beneficial for the AAM loss, but
the metric learning approach itself brings an improvement even without it. Finally, adding the GAT branch to
ML-Decoder doesn’t increase the accuracy, indicating that additional information, coming from GAT, gives no new cues
to ML-Decoder.

\begin{table}[h!]
  \caption{Algorithm's components contribution.}
  \label{tab:comp_ablation}
  \centering
  \begin{tabular}{l|c|c|c|c}
    Method & P-VOC & COCO & N-WIDE & VG500 \\
    \hline
    baseline & 93.58 & 85.90 & 63.85 & 37.85 \\
    + SAM & 94.00 & 86.75 & 65.20 & 38.50 \\
    + OI wghts & 94.24 & 87.05 & 65.54 & 38.84 \\
    + MLD & 95.54 & 88.30 & 67.07 & 41.20\\
    + AM loss$^*$ & 95.80 & 88.60 & 67.30 & 41.90 \\
    + AAM & 95.86 & 88.75 & 67.60 & 42.00 \\
    + GAT & 95.85 & 88.70 & 67.20 & - \\
    \multicolumn{5}{l}{\footnotesize{$^*$ AAM with $\gamma+=\gamma-=0$}}
  \end{tabular}
\end{table}

\section{\uppercase{Conclusion}}
In this work, we revisited two popular approaches to multilabel classification:
transformer-based heads and labels graph branches. We refined the performance of these approaches
by applying our training strategy with the modern bag of tricks and introducing a novel loss for multilabel classification called AAM.
The loss combines properties of the ASL loss and metric learning approach and allows
achieving competitive results on popular multilabel benchmarks. Although we demonstrated that graph
branches perform very close to transformer-based heads, there is one major drawback of the graph-based method: it relies on
labels’ representations provided by a language model. The direction of  future work could be developing an approach which
would build a label relations graph relying on representations extracted directly from images, not involving potentially
meaningless label names.

\bibliographystyle{apalike}
{\small
\bibliography{refs}}

\section{\uppercase{Supplementary materials}}

\subsection{Extended ablation study}
\subsubsection{Hyperparameters impact}
\label{sec:hyperparameters}

\begin{figure}[ht!]
  \centering
  \subfloat[COCO]{{\includegraphics[width=0.4\textwidth]{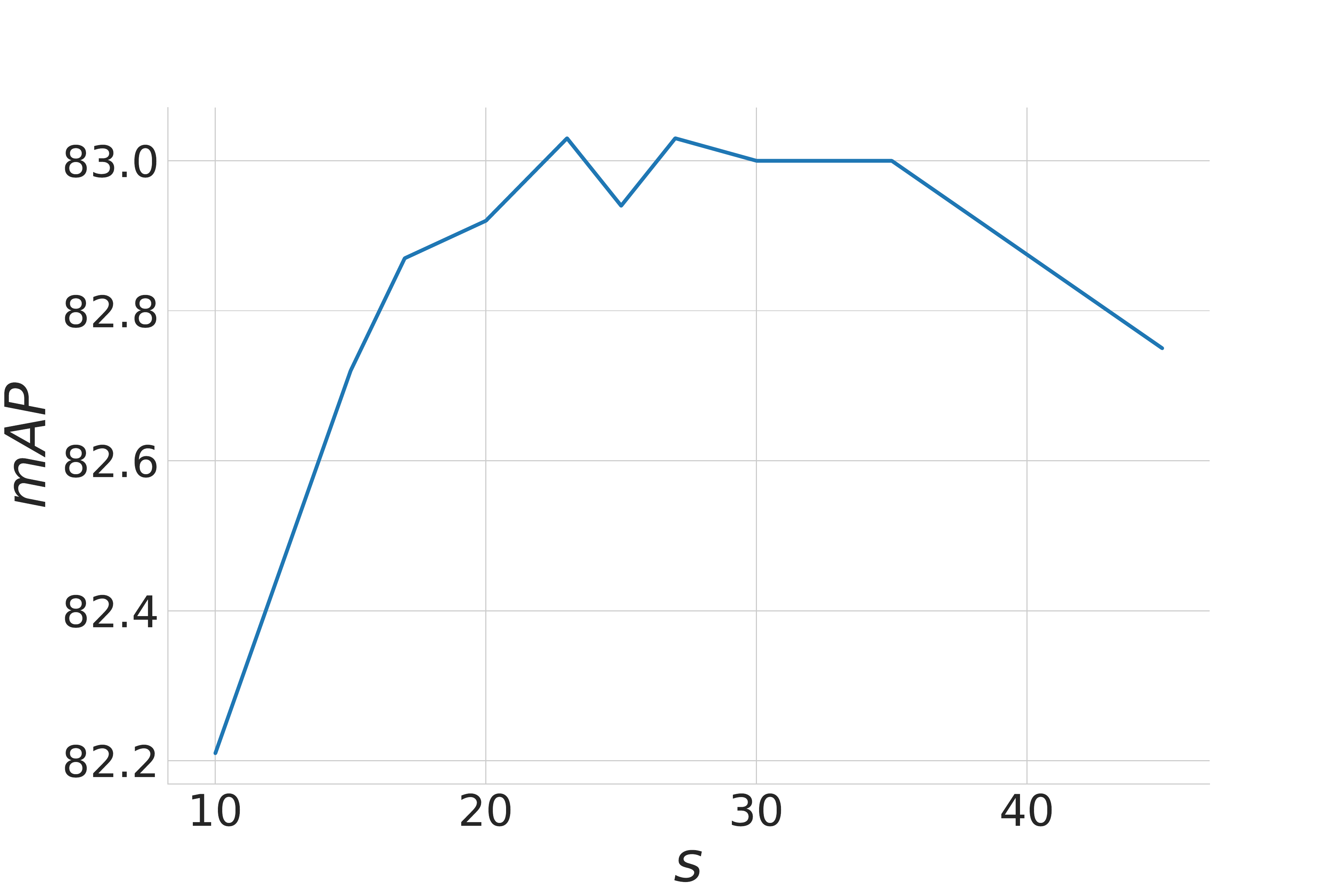} }}
  \qquad
  \subfloat[VOC]{{\includegraphics[width=0.4\textwidth]{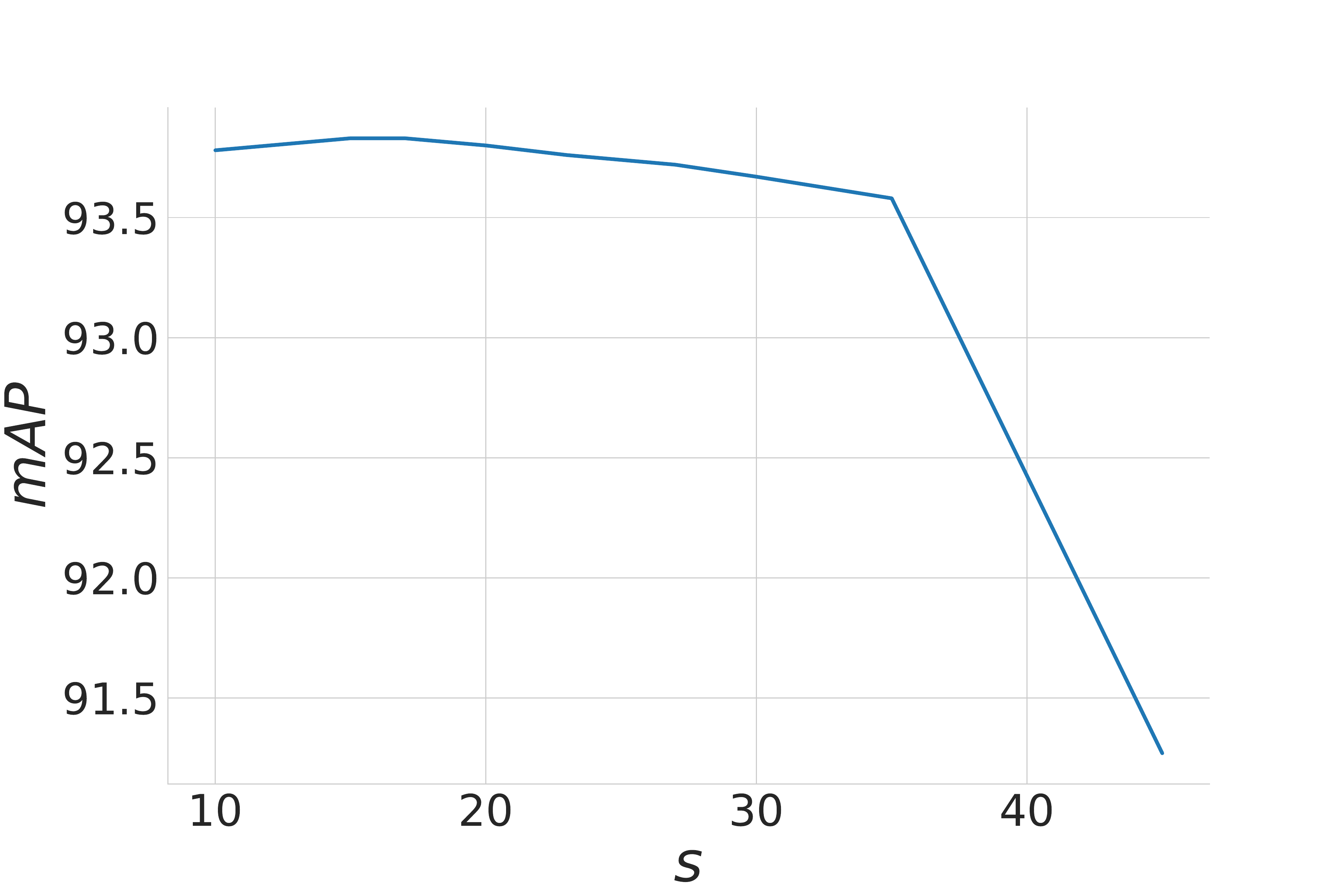} }}
  \caption{The performance curves of EfficientNetV2-s+MLD+AAM model with different values of the $s$ parameter.
  Margin parameter in AAM loss is set to zero. Training resolution is 224x224.}
  \label{img:abl_scale}
\end{figure}

Figure \ref{img:abl_scale} shows the scale parameter $s$ impact on the training pipeline.
We fix margin parameter at 0.0 value. We use our full training pipeline with AAM loss and
ML-Decoder on 224x224 resolution for faster training.
We can conclude that the value of $s$ is correlated with the number of classes.
For a small number of classes, the optimal value lies in the range 10-20.
Otherwise, the 20-30 range will be a good choice.

\begin{figure}[t!]
  \centering
  \subfloat[VOC]{{\includegraphics[width=0.4\textwidth]{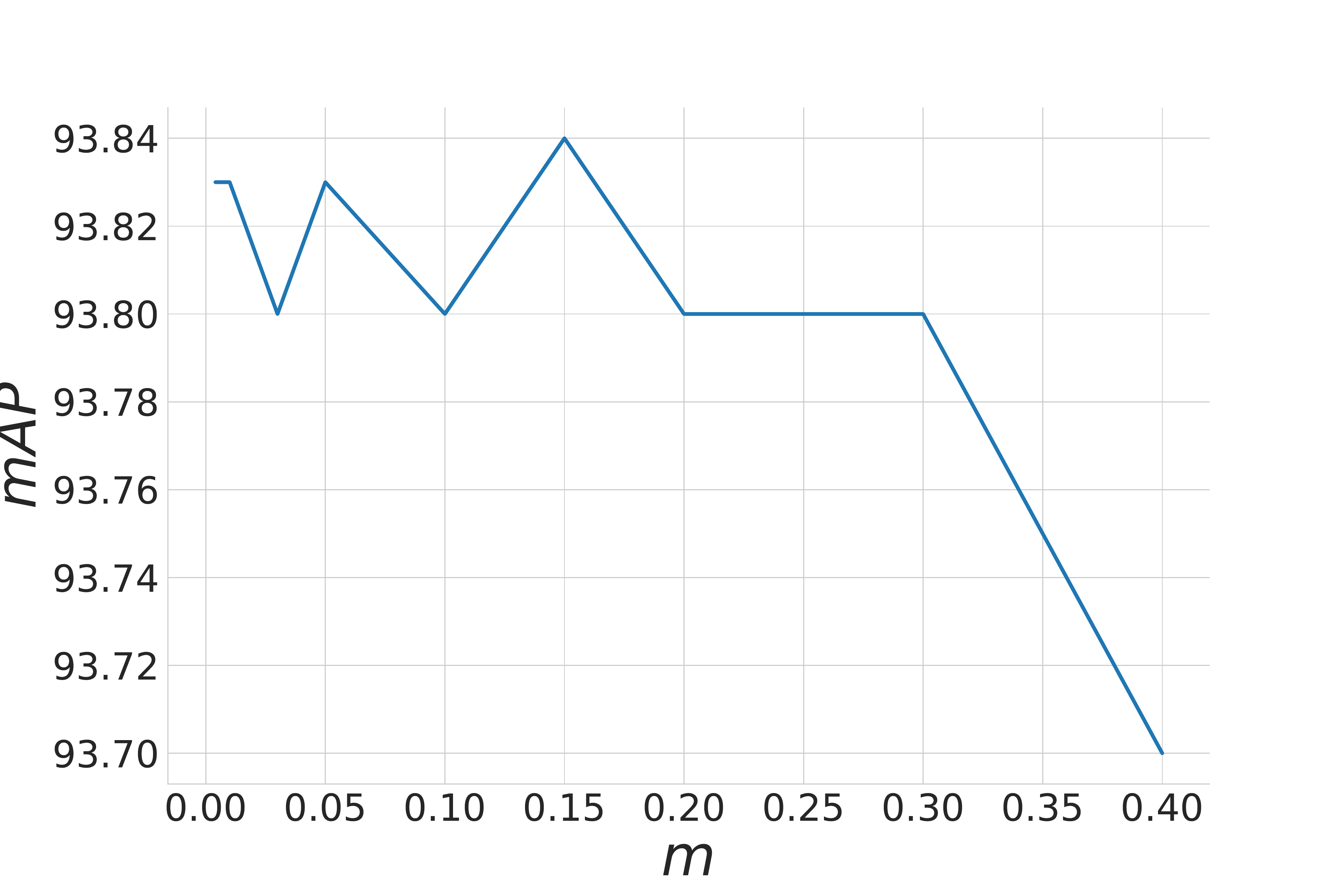} }}
  \qquad
  \subfloat[COCO]{{\includegraphics[width=0.4\textwidth]{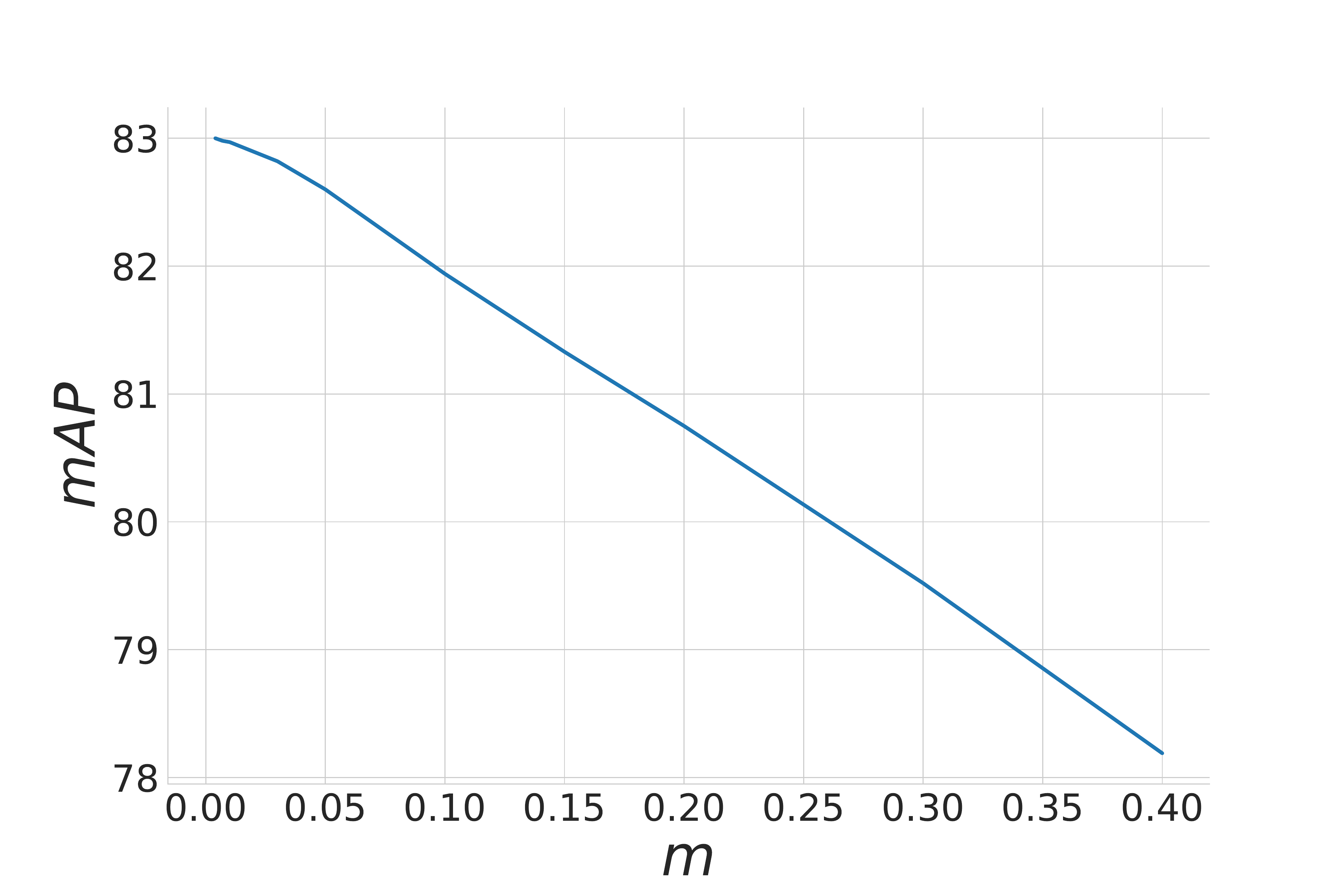} }}
  \caption{The performance curves of EfficientNetV2-s+MLD+AAM model with different values of the $m$ parameter.
  We fixed other hyperparameters for each dataset for all experiments. Training resolution is 224x224.}
  \label{img:abl_margin}
\end{figure}

Figure \ref{img:abl_margin} shows margin influence from AAM loss.
We see that this parameter brings unnecessary complexity to the choice of extra hyperparameter.
We can exclude this parameter and simplify applying of the loss function.

Table \ref{tab:abl_gamma} shows the impact of the different asymmetry parameters values $\gamma^+$ and $\gamma^-$.
As the authors of ASL \cite{Baruch2021AsymmetricLF} state in their work, a larger $\gamma^-$
is required to handle a larger positive-negative imbalance.
They also set $\gamma^+$ to 0 for all experiments, but we observe that there are datasets
where non-zero $\gamma^+$ in AAM loss could bring an improvement as well.

We also notice that in the case of AAM loss, we need just 0-2 values ranges,
as we have additional global loss scale parameter $\frac{1-k}{s}$ in (\ref{eq:aam}).

\begin{table*}[h]
  \caption{Asymmetry influence on the training pipeline.}
  \label{tab:abl_gamma}
  \centering
  \begin{tabular}{l|c|c|c|c}
    Gamma & VOC-2007 & COCO & NUS-WIDE & VG500 \\
    \hline
    $\gamma^-$ = 0, $\gamma^+$ = 0 & 95.80  & 88.60 & 67.20 & 41.90 \\
    $\gamma^-$ = 1, $\gamma^+$ = 0 & 95.80  & \textbf{88.75} & 67.30 & \textbf{42.00} \\
    $\gamma^-$ = 2, $\gamma^+$ = 0 & 95.77  & 88.68  & 67.47 & 41.90 \\
    $\gamma^-$ = 3, $\gamma^+$ = 0 & 95.78  & 88.62  & 67.27 & 41.80  \\
    $\gamma^-$ = 2, $\gamma^+$ = 1 & \textbf{95.86} & 88.64 & \textbf{67.60} & 41.95 \\
    $\gamma^-$ = 3, $\gamma^+$ = 1 & 95.80 & 88.52  & \textbf{67.60} & 41.60 \\
    \hline
  \end{tabular}
\end{table*}

\subsubsection{Graph attention branch and transformer head comparison}

\begin{table*}[t!]
  \caption{Comparison of GAN vs GCN method and ML Decoder.}
  \label{tab:gcn_ablation}
  \centering
  \begin{tabular}{l|c|c|c}
    Step & VOC-2007 & COCO & NUS-WIDE \\
    \hline
    ML-GCN EfficientNetV2-s & 95.25 & 87.50 & 66.30 \\
    GAN re-weighting EfficientNetV2-s & \textbf{96.00} & 87.70 & 66.85 \\
    EfficientNetV2-s + MLD & 95.86 & \textbf{88.75} & \textbf{67.60} \\
    GAN re-weighting EfficientNetV2-s + MLD & 95.85 & 88.70 & 67.20 \\
    \hline
  \end{tabular}
\end{table*}

\begin{table*}[t!]
  \caption{Comparison of graph-based method and transformer based in low resolution setup.}
  \label{tab:gcn_ablation_224}
  \centering
  \begin{tabular}{l|c|c|c}
    Step & VOC-2007 & COCO & NUS-WIDE \\
    \hline
    EfficientNetV2-s & 92.85 & 82.25 & 65.60\\
    GAN re-weighting EfficientNetV2-s & 93.15 & 82.55 & 65.87 \\
    EfficientNetV2-s + MLD & \textbf{93.77} & \textbf{83.03} & \textbf{66.74} \\
    GAN re-weighting EfficientNetV2-s + MLD & 93.71 & \textbf{83.03} & 66.60 \\
    \hline
  \end{tabular}
\end{table*}

In this section we compare our proposed new re-weighting scheme and the GCN method.
If we refer to ML-GCN work \cite{Chen2019MultiLabelIR}, we see that the training strategy the authors chose for their method is rather simple.
We argue that if we add a modern bag of training tricks to the method, the accuracy potentially could increase.
We train backbone with ML-GCN head as described in \cite{Chen2019MultiLabelIR},
but with our training strategy as described in Section \ref{sec:training_strategy}. Table \ref{tab:gcn_ablation}
shows that GAN with a re-weighting scheme is more sophisticated in incorporating inter-label correlation.
At the same time, the global attention ability of the transformer is sufficient for seeking internal dependencies.
Even if we combine two methods and add to the model apriori information about the conditional probability of appearing labels,
we will not obtain better quality.

The next point is to check the ability of the transformer and GAN model to handle case of the high level of label noise.
A high level of label noise can be achieved by simply reducing the image resolution.
Most of the small objects (especially in COCO and NUS-WIDE datasets) will disappear because of resize artifacts.

Table \ref{tab:gcn_ablation_224} shows the obtained results on all data with resolution 224x224.
We also present the results of the model without both ML Decoder and GAN. We can conclude,
that despite the fact that GAN branch is sufficient to help the ordinary model to
obtain global information on label distribution and co-occurrence,
the transformer-based head can derive this information implicitly from the CNN features in a more efficient way.
Again, adding GAN as an additional branch doesn't improve the results.

\subsection{Confidence threshold tuning}

Conventionally, threshold a value equals to $0.5$ is used for calculating OP, OR, OF1 and
CP, CR, CF1. But to apply a classification model in the wild under the variety of input data
and presented classes, one need to estimate per-class confidence thresholds. When we don't have
an access to the target real-world data, we can at least take into account per-class threshold
variance, which arises because of training data distribution and loss function modifications.
To do that, we propose finding per class thresholds by maximizing F1 score on each class
via grid search on the train subset. Then we can evaluate the obtained thresholds on the
validation subset and check if the precision-recall balance was improved.

\begin{table}
\caption{Results of confidence thresholds calibration for EfficientNetV2-s+MLD model trained with AAM.}
\label{tab:thresholding}
\centering
\begin{tabular}{l|c|c|c|c}
    Dataset & OF1 & OF1-adapt & CF1 & CF1-adapt \\ \hline
    P-VOC & 91.14 & 91.94 & 92.60 & 93.21 \\
    COCO & 82.86 & 84.39 & 85.03 & 86.06 \\
    N-WIDE & 71.88 & 75.52 & 72.90 & 75.04\\
    VG500 & 56.04 & 54.09 & 55.15 & 53.07\\
\end{tabular}
\end{table}

In Table \ref{tab:thresholding} the results of thresholds tuning for EfficientNetV2-s model trained with
AAM loss on several datasets are shown. Precision-recall balance was improved by the mentioned
procedure under train-validation distribution shift conditions for all of the considered datasets excepting
Visual Genome 500. There thresholds tuning on train slightly decreased both OF1 and CF1 scores.
This fact indicates that VG500 has a big distribution gap between the train and validation subsets, which
the model can't handle (mAP $< 50\%$). Thus, if the trained model has low accuracy, estimating confidence
threshold on the training subset may not work as expected.

\end{document}